\PassOptionsToPackage{table}{xcolor}
\documentclass{arxiv_style/arxiv_singlecolumn}
\usepackage{arxiv_style/rubricarmor}

\title{Rubric\elegantArmor{}: Adversarial Evolution Improves LLM-Based Rubric Generation}

\author[1]{Haocheng~Yang}
\author[2]{Yuchao~Zhang}
\author[3]{Licheng~Pan}
\author[4]{Jiajun~Fan}
\author[5]{Maolin~Wang}
\author[6]{Kangning~Zhang}
\author[6]{Shuai~Shao}
\author[7]{Shijian~Wang}
\author[8]{Yuan~Lu}
\author[9]{Chunyuan~Zheng}
\author[10*]{Hao~Wang}

\affiliation[1]{National University of Singapore}
\affiliation[2]{Beijing University of Chemical Technology}
\affiliation[3]{Zhejiang University}
\affiliation[4]{University of Illinois at Urbana-Champaign}
\affiliation[5]{City University of Hongkong}
\affiliation[6]{Shanghai Jiaotong University}
\affiliation[7]{Southeast University}
\affiliation[8]{Xiaohongshu}
\affiliation[9]{Peking University}
\affiliation[10]{MBZUAI}

\contribution[*]{Corresponding author}
\metadata[Contact]{\href{mailto:haocheng_yang@u.nus.edu}{\textcolor{black}{haocheng\_yang@u.nus.edu}}, \email{Ho-ward@outlook.com}}
\metadata[Code]{\codeurl}
\hypersetup{
  pdftitle={RubricArmor: Adversarial Evolution Improves LLM-Based Rubric Generation},
  pdfauthor={Haocheng Yang, Yuchao Zhang, Licheng Pan, Jiajun Fan, Maolin Wang, Kangning Zhang, Shuai Shao, Shijian Wang, Yuan Lu, Chunyuan Zheng, Hao Wang}
}

\abstract{Rubric-based reinforcement learning (RL) provides interpretable rewards for aligning large language models (LLMs) by evaluating responses against query-specific evaluation criteria. To construct rubrics at scale, a straightforward approach to LLM-based rubric generation is to prompt an LLM to generate a rubric directly from the query. However, rubrics directly generated by LLMs are vulnerable to reward hacking, since omitted or underspecified criteria allow the policy to obtain high rubric rewards with low-quality responses. Existing LLM-based rubric generation methods improve the granularity and coverage of the generated criteria but do not proactively guard against reward hacking. To address this limitation, we propose RubricArmor, an adversarial framework that exposes and mitigates potential reward hacking at the rubric generation stage before it occurs in subsequent RL. Specifically, RubricArmor performs adversarial evolution, in which an attack step and a repair step alternate over multiple rounds. The attack step simulates the reward hacking of the policy by constructing adversarial responses that satisfy the current rubric but fail to properly complete the task. The repair step then revises the rubric to detect the response defects exposed by the attack step while preserving other valid criteria. Extensive experiments demonstrate that RubricArmor outperforms competitive rubric generation baselines and translates into more effective downstream rubric-based RL.}

\begin{document}
\maketitle

\section{Introduction}
\label{sec:introduction}

Reinforcement learning from human feedback (RLHF) has emerged as a cornerstone technique for aligning large language models (LLMs) with human preferences~\citep{ouyang2022training}. Among RLHF paradigms, rubric-based RL provides interpretable and task-specific rewards~\citep{gunjal2025rubrics,he2025rifl}. Rubric-based RL typically adopts a two-stage pipeline: first, a rubric composed of multiple evaluation criteria is constructed for each query; second, the policy is optimized by RL to maximize the rubric reward, i.e., the fraction of criteria satisfied by the response. Consequently, constructing accurate and comprehensive rubrics is essential for rubric-based RL, as the rubric determines which behaviors are rewarded and thus reinforced.

Since expert annotation of rubrics is costly and difficult to scale~\citep{arora2025healthbench}, recent work has developed LLM-based rubric generation, which uses LLMs to infer the requirements of an ideal response and express them as verifiable criteria. A straightforward approach is to prompt an LLM to generate a rubric directly from the query and use the resulting criteria as rewards for policy optimization. However, \textbf{rubrics directly generated by LLMs are vulnerable to reward hacking}. Specifically, generated rubrics may omit essential criteria or leave criteria underspecified. When an essential criterion is omitted, a response can violate the corresponding quality requirement without losing reward. When a criterion is underspecified, a response can satisfy the stated conditions and receive credit without meeting the intended quality standard. Since the policy actively searches for behaviors that maximize the rubric reward, the policy can discover and repeatedly produce high-scoring but low-quality responses during RL~\citep{mahmoud2026hacking}.

Existing LLM-based rubric generation methods improve upon the straightforward approach but do not fully address the vulnerability to reward hacking. One line of work constructs finer-grained evaluation checklists~\citep{cook2024ticking,wei2025rocketeval}, without testing whether a low-quality response can still satisfy every checklist item. Another line of work progressively refines the generated criteria with reference responses or preference comparisons~\citep{zhang2026chasing,shen2026rethinking,li2026rubrichub,rezaei2025online}, but the refinement relies on available responses rather than responses that actively exploit the rubric.

To address this limitation, we propose \textbf{RubricArmor}, an adversarial framework for LLM-based rubric generation. The core idea is to expose potential reward hacking at the rubric generation stage before it occurs in subsequent RL, and to mitigate it through adversarial evolution, an iterative attack--repair loop. The attack step simulates the reward hacking that the policy may perform by constructing adversarial responses that satisfy the current rubric but fail to properly complete the task, thereby exposing the response defects that the rubric fails to detect. The repair step then revises the rubric to detect the exposed defects while preserving other valid criteria. Through repeated rounds of attack and repair, the rubric progressively covers more boundary cases and implicit requirements and converges to a robust and accurate rubric, which both resists potential reward hacking and guides policies toward high-quality responses.

The main contributions of this work are summarized as follows. \ding{182} \textbf{We identify a critical limitation of LLM-based rubric generation:} rubrics directly generated by LLMs are vulnerable to reward hacking, since omitted or underspecified criteria allow high-scoring but low-quality responses. \ding{183} \textbf{We propose RubricArmor to expose and mitigate potential reward hacking before RL.} RubricArmor iteratively simulates the reward hacking of the policy with adversarial responses and repairs the rubric to detect the exposed response defects. \ding{184} \textbf{We conduct comprehensive experiments to evaluate the efficacy of RubricArmor}, where RubricArmor exceeds the strongest baseline by 3.12 points in macro accuracy on nine preference evaluation sets and achieves the best downstream RL performance in 13 of 15 settings across three policy models.

\section{Preliminaries}
\label{sec:preliminaries}

\subsection{Rubric-Based Reinforcement Learning}
\label{sec:rubric_based_rl}

Reinforcement learning from human feedback (RLHF) aligns language model responses with human preferences~\citep{ouyang2022training}.
Suppose $x$ denotes a query, $y$ a response, and $\pi_\theta$ a language model policy parameterized by $\theta$.
Given a scalar reward $s(x,y)$, policy optimization maximizes its expectation over a query distribution $\mathcal{D}$:
\begin{equation}
    \max_{\theta}\;
    \mathbb{E}_{x\sim\mathcal{D},\;y\sim\pi_\theta(\cdot\mid x)}
    \bigl[s(x,y)\bigr].
    \label{eq:policy_objective}
\end{equation}
Common optimization algorithms include proximal policy optimization (PPO)~\citep{schulman2017proximal}, group relative policy optimization (GRPO)~\citep{shao2024deepseekmath}, and group sequence policy optimization (GSPO)~\citep{zheng2025gspo}.

Rubric-based RL provides interpretable, task-specific rewards by evaluating responses against explicit criteria and aggregating the resulting judgments~\citep{gunjal2025rubrics,li2026rubrichub,li2026ares}.
Formally, let $\mathcal{R}=\{c_1,\ldots,c_m\}$ denote a rubric comprising $m$ criteria for query $x$, each specifying a checkable requirement.
An LLM-based verifier returns $v(x,y,c_i)\in\{0,1\}$ for each criterion, where $1$ denotes \textsc{Pass} and $0$ denotes \textsc{Fail}.
We use equal weights, giving the fraction of satisfied criteria as the reward:
\begin{equation}
    s(x,y)
    = \frac{1}{m}\sum_{i=1}^{m}v(x,y,c_i).
    \label{eq:rubric_reward}
\end{equation}

\subsection{LLM-Based Rubric Generation}
\label{sec:problem_definition}

Rubric generation aims to construct evaluation criteria tailored to each query.
Some approaches rely on expert annotation~\citep{arora2025healthbench} or train rubric generators on annotated data~\citep{liu2025openrubrics,xu2026alternating,jiang2026rubricarrow}, both of which are costly and difficult to scale.
To address this limitation, \textbf{LLM-based rubric generation} prompts LLMs to infer the requirements of an ideal response and express them as verifiable criteria, without training a rubric generator.

One line of work prompts LLMs to translate query requirements directly into evaluation checklists~\citep{cook2024ticking,wei2025rocketeval}.
Another line of work progressively refines and expands the generated criteria to improve their granularity and coverage~\citep{zhang2026chasing,shen2026rethinking,gao2026qworld,li2026rubrichub}.

In this work, we aim to improve the quality of LLM-based rubric generation.
Given a query $x$, our goal is to construct a rubric $\mathcal{R}$ that accurately and comprehensively captures the task requirements, so that $s(x,y;\mathcal{R})$ reflects the quality of response $y$.

\section{Methodology}
\label{sec:methodology}

\subsection{Motivation}
\label{sec:motivation}

LLM-based rubric generation is central to constructing high-quality rubrics across diverse queries. By translating query requirements into explicit evaluation criteria, it determines which response properties are rewarded during rubric-based RL. Generating accurate and comprehensive criteria is therefore crucial for providing reliable rewards and guiding policies toward better task performance.

A straightforward approach to LLM-based rubric generation is to prompt an LLM to generate a rubric directly from the query and use the resulting criteria to compute rewards for policy optimization. \textbf{However, rubrics directly generated by LLMs are vulnerable to reward hacking.} Specifically, generated rubrics may omit  criteria essential for satisfying the user query or specify them incompletely. When an essential criterion is omitted, a response can violate the corresponding quality requirement without losing reward. When a criterion is underspecified, a response can satisfy its stated conditions and receive credit without meeting the intended quality standard. In both cases, rubric rewards can remain high despite response defects, allowing policy optimization to reinforce low-quality responses and degrade downstream performance.

\begin{figure}[t]
    \centering
    \captionsetup[subfigure]{skip=3pt}
    \newsavebox{\rubriccaseimage}
    \sbox{\rubriccaseimage}{\includegraphics[width=0.48\linewidth,pagebox=cropbox]{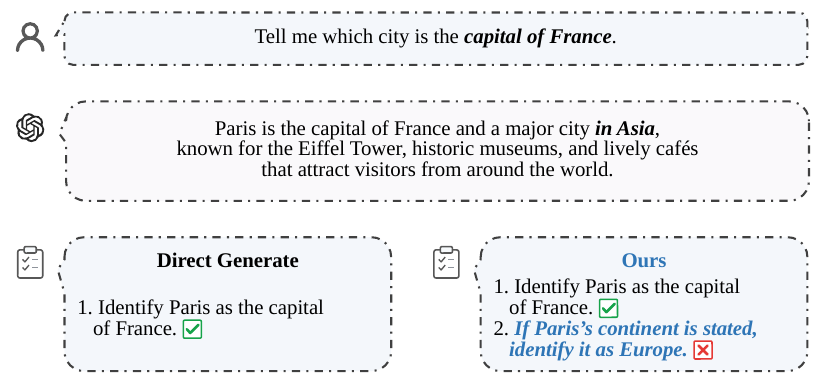}}
    \begin{minipage}[t]{0.48\linewidth}
        \vspace{0pt}
        \centering
        \subcaptionbox{Omit essential criteria.\label{fig:rubric_hacking_coverage}}{\usebox{\rubriccaseimage}}
    \end{minipage}\hfill
    \begin{minipage}[t]{0.015\linewidth}
        \vspace{0pt}
        \centering
        \resizebox{!}{\ht\rubriccaseimage}{%
            \begin{tikzpicture}
                \draw[dash pattern=on 14pt off 10.5pt, line width=2pt] (0,0) -- (0,8);
            \end{tikzpicture}%
        }
    \end{minipage}\hfill
    \begin{minipage}[t]{0.48\linewidth}
        \vspace{0pt}
        \centering
        \subcaptionbox{Underspecified criteria.\label{fig:rubric_hacking_specification}}{\includegraphics[width=\linewidth,pagebox=cropbox]{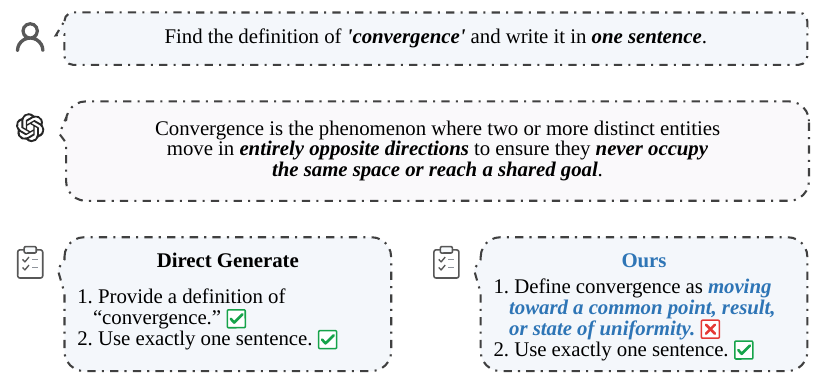}}
    \end{minipage}
    \caption{\textbf{Two examples of rubric reward hacking.} (a) Omitting essential criteria leaves an additional factual error unchecked. (b) Underspecified criteria allow an incorrect response to pass. The refined rubrics (Ours) capture both defects.}
    \label{fig:rubric_hacking_case}
\end{figure}

\textbf{Case study.} To further illustrate the above limitation, we present two typical scenarios in which a rubric generated directly by an LLM is exploited by a poor-quality response to obtain full reward. In Figure~\ref{fig:rubric_hacking_case}(a), the response correctly identifies Paris as the capital of France but incorrectly places it in Asia. The directly generated rubric checks only the capital, leaving the additional factual error undetected. In Figure~\ref{fig:rubric_hacking_case}(b), the directly generated definition criterion leaves the meaning of convergence under-specified, allowing the response to receive full reward despite describing the concept wrong. The refined rubrics (Ours) capture these errors by adding the missing factual check and making the definition explicit respectively. Therefore, these cases demonstrate that rubrics directly generated by LLMs are vulnerable to reward hacking.

Given the importance of LLM-based rubric generation and the vulnerability of generated rubrics to reward hacking, we investigate three key questions. How can we expose and mitigate reward hacking during LLM-based rubric generation? Does mitigating reward hacking during rubric generation improve the quality of generated rubrics? Do improvements in rubric quality translate into better downstream RL performance?
\subsection{Adversarial Rubric Generation}
\label{sec:adversarial_generation}

To overcome the vulnerability of rubrics directly generated by LLMs to reward hacking, we propose RubricArmor, an adversarial framework for LLM-based rubric generation. The core idea is to expose potential reward hacking at the rubric generation stage before it occurs in subsequent RL, and to mitigate it through active attack and repair.

To achieve this, RubricArmor runs an iterative attack--repair loop. The attack step simulates the reward hacking that the policy may perform in subsequent RL. Specifically, the attack step constructs adversarial responses that satisfy the current rubric but fail to properly complete the task. These adversarial responses expose the response defects that the rubric fails to detect. The repair step then analyzes the adversarial responses and repairs the rubric to detect the exposed defects. Through repeated rounds of attack and repair, the rubric progressively covers more boundary cases and implicit requirements and finally converges to a robust and accurate rubric. The converged rubric both resists potential reward hacking and guides policies toward high-quality responses.

\begin{figure}[t]
    \centering
    \includegraphics[width=\textwidth]{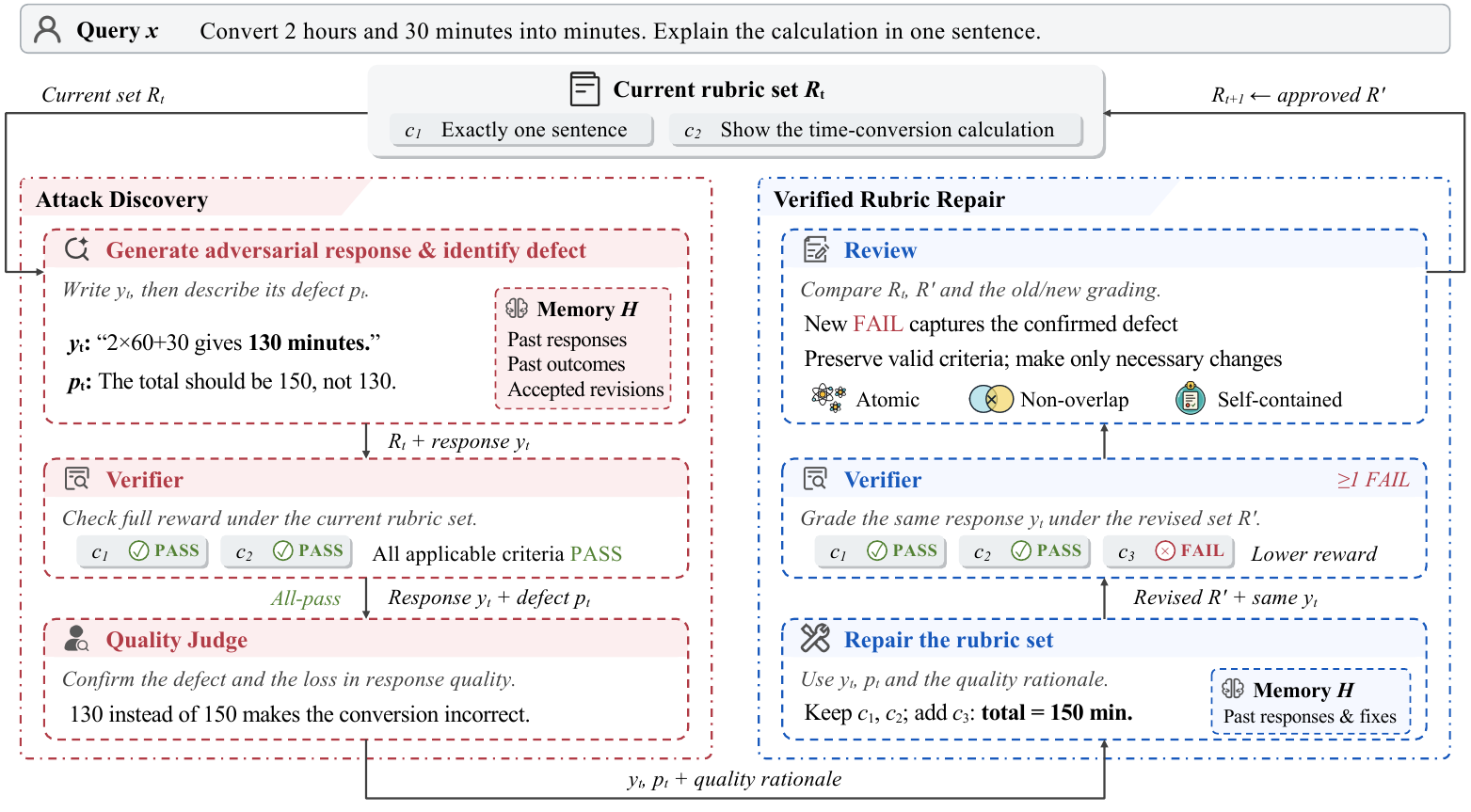}
    \caption{\textbf{Overview of a RubricArmor evolution round.} Attack Discovery constructs adversarial responses that satisfy the current rubric and independently confirms their defects and the resulting loss in response quality. Verified Rubric Repair revises the rubric to detect the confirmed defects, verifies that the adversarial response receives a lower reward, and reviews the revised criteria. Each accepted rubric becomes the target of subsequent attacks.}
    \label{fig:arme_overview}
\end{figure}

To this end, we implement RubricArmor as a fully automated multi-agent system with two modules, \textbf{Attack Discovery} and \textbf{Verified Rubric Repair}, as illustrated in Figure~\ref{fig:arme_overview}. We introduce the two modules in detail as follows.

\subsubsection{Attack Discovery}
\label{sec:attack_discovery}

Attack Discovery aims to expose the reward hacking that the policy may perform under the current rubric. The module consists of an attacker, the verifier, and an independent quality judge. Suppose $\mathcal{R}_t$ denotes the rubric for query $x$ at round $t$, and $H$ denotes a memory that records past adversarial responses, their outcomes, and accepted revisions. Guided by $H$, the attacker constructs an adversarial response $y_t$ and identifies a potential defect $p_t$ in the response.

An attack is successful when two conditions hold: $y_t$ satisfies the current rubric, and $y_t$ contains a defect $p_t$ that substantially reduces response quality. Attack Discovery checks the two conditions in sequence. First, the verifier grades $y_t$ against each criterion to determine whether $y_t$ receives full reward. Then, for a response receiving full reward, the quality judge checks whether $p_t$ is present in $y_t$ and whether $p_t$ substantially reduces response quality given the query and its context. Let $q_t=1$ indicate that the quality judge confirms the second condition. Formally, a successful attack satisfies
\begin{equation}
    s(x,y_t;\mathcal{R}_t)=1
    \quad\text{and}\quad q_t=1.
    \label{eq:attack_acceptance}
\end{equation}
Each successful attack thus exposes reward hacking under $\mathcal{R}_t$ and provides concrete evidence for Verified Rubric Repair.

\subsubsection{Verified Rubric Repair}
\label{sec:rubric_repair}

Verified Rubric Repair aims to revise the rubric to detect the confirmed defect $p_t$ in $y_t$, while preserving other valid criteria. The module consists of a repair agent, the verifier, and a review agent. Given $y_t$ and $p_t$, the repair agent revises $\mathcal{R}_t$ into a revised rubric $\mathcal{R}'$. To preserve earlier fixes when addressing the new defect, the repair agent also consults the past adversarial responses and accepted revisions in $H$.

When revising the rubric, the repair agent requires each criterion to be atomic, self-contained, and non-overlapping \citep{shen2026rethinking}, so that individual criteria can be evaluated and revised independently. An atomic criterion checks one independently judgeable requirement. A self-contained criterion states the facts, values, or relations needed to evaluate the response. Non-overlapping criteria check distinct requirements, avoiding repeated reward for the same response property. In particular, the three properties aim for criteria that can be violated individually while the remaining criteria are satisfied.

A revision is accepted when two conditions hold: the revised rubric assigns a lower reward to $y_t$, and the revised criteria pass the review. RubricArmor checks the two conditions in sequence. First, the verifier grades $y_t$ under $\mathcal{R}'$ to determine whether $y_t$ fails at least one criterion. Then, the review agent checks that the revised criteria detect the confirmed defect and that other valid criteria and the three properties are preserved. Let $b_t=1$ indicate that the review agent approves the revision. Formally, RubricArmor accepts the revised rubric when
\begin{equation}
    s(x,y_t;\mathcal{R}')<1
    \quad\text{and}\quad b_t=1.
    \label{eq:repair_acceptance}
\end{equation}
Each accepted revision thus mitigates the exposed reward hacking while preserving other valid criteria.

\subsection{The workflow of RubricArmor.}
\label{sec:arme_workflow}

Algorithm~\ref{alg:arme} summarizes the complete workflow of RubricArmor. An initializer first constructs $\mathcal{R}_0$ from query $x$, with the memory $H$ initially empty. Attack Discovery and Verified Rubric Repair then alternate, with each accepted revision becoming the rubric for the next round. Meanwhile, $H$ accumulates the adversarial responses, their outcomes, and the accepted revisions to guide subsequent rounds. When no revision is accepted, Verified Rubric Repair returns the current rubric.

Evolution stops after at most $T$ rounds or $P$ consecutive rounds without an accepted revision. The last accepted rubric is returned to compute rewards for downstream policy optimization.

\begin{algorithm}[t]
    \caption{The workflow of RubricArmor.}
    \label{alg:arme}
    \begin{algorithmic}[1]
        \Require Query $x$; maximum rounds $T$; patience $P$
        \Ensure Evolved rubric $\mathcal{R}$
        \State $\mathcal{R}\gets\Call{Initialize}{x}$; $H\gets\emptyset$; $u\gets0$
        \For{$t=1,\ldots,T$}
            \State $(y_t,p_t,H)\gets\Call{AttackDiscovery}{x,\mathcal{R},H}$; $\mathcal{R}'\gets\mathcal{R}$
            \If{attack succeeds}
                \State $(\mathcal{R}',H)\gets\Call{VerifiedRubricRepair}{x,\mathcal{R},y_t,p_t,H}$
            \EndIf
            \State $u\gets0$ \textbf{if} $\mathcal{R}'\ne\mathcal{R}$ \textbf{else} $u+1$
            \State $\mathcal{R}\gets\mathcal{R}'$
            \State \textbf{if} $u\ge P$ \textbf{then break}
        \EndFor
        \State \Return $\mathcal{R}$
    \end{algorithmic}
\end{algorithm}

\section{Experiments}
\label{sec:experiments}

In this section, we empirically validate the efficacy of RubricArmor, centered on five research questions (RQs) as follows:
\begin{itemize}[leftmargin=*]
    \item \textbf{RQ1}: \textit{Does RubricArmor perform well?} In Section~\ref{sec:overall_performance}, we compare RubricArmor against competitive baselines on nine preference evaluation sets.
    \item \textbf{RQ2}: \textit{Why does RubricArmor work?} In Section~\ref{sec:ablation}, we perform an ablation study on response construction and the adversarial objective.
    \item \textbf{RQ3}: \textit{Does RubricArmor generalize across rubric generator backbones?} In Section~\ref{sec:generalization}, we evaluate RubricArmor with backbones of different families and sizes.
    \item \textbf{RQ4}: \textit{Is RubricArmor sensitive to hyperparameters?} In Section~\ref{sec:evolution_dynamics}, we analyze the sensitivity of RubricArmor to the maximum number of evolution rounds.
    \item \textbf{RQ5}: \textit{Does RubricArmor improve downstream RL?} In Section~\ref{sec:downstream_rl}, we validate its practical utility to rubric-based RL on downstream benchmarks.
\end{itemize}

\begin{table}[!ht]
\centering
\caption{Pairwise preference accuracy (\%) across nine evaluation sets from three benchmark suites.}
\label{tab:main_results}
\scriptsize
\setlength{\tabcolsep}{3pt}
\renewcommand{\arraystretch}{1.12}
\resizebox{\textwidth}{!}{%
\begin{tabular}{l*{10}{c}}
\toprule
& \multicolumn{4}{c}{\textbf{JudgeBench}}
& \multicolumn{4}{c}{\textbf{RM-Bench}}
& \textbf{RewardBench} & \textbf{Macro} \\
\cmidrule(lr){2-5}\cmidrule(lr){6-9}\cmidrule(lr){10-10}\cmidrule(lr){11-11}
\textbf{Method} & Knowledge & Reasoning & Math & Coding
& Chat & Math & Code & Safety & Chat-Hard & Avg. \\
\midrule
\multicolumn{11}{l}{\cellcolor{tablegray}\textbf{\textit{LLM-based rubric generation}}} \\
Direct & 62.01 & 53.69 & 70.56 & 69.18 & 64.38 & 77.59 & 69.03 & 89.91 & 74.67 & 70.11 \\
TICK & 65.75 & 58.39 & 73.33 & 67.12 & 66.67 & 72.13 & 69.76 & 90.61 & 75.22 & 71.00 \\
RocketEval & 70.78 & 63.42 & 74.44 & 68.49 & 65.59 & 82.13 & 70.47 & 88.23 & 73.14 & 72.97 \\
ChasingTail & 72.40 & \textbf{68.12} & 76.67 & 67.12 & 66.45 & 84.58 & 69.62 & 88.38 & 72.37 & 73.97 \\
RRD-Uniform & 72.56 & 67.79 & 77.78 & 70.55 & 59.95 & 83.67 & 66.06 & 80.91 & 69.74 & 72.11 \\
\hdashline
\multicolumn{11}{l}{\cellcolor{tablegray}\textbf{\textit{Trained rubric generators}}} \\
Rubric-ARM-8B & 58.28 & 58.72 & 70.00 & 71.92 & 68.13 & 67.78 & 68.86 & 74.14 & 78.29 & 68.46 \\
Rubric-RM-8B & 57.79 & 60.40 & 72.78 & 68.49 & 66.24 & 67.19 & 68.98 & 74.77 & 74.89 & 67.95 \\
Rubric-ARROW-8B & 53.08 & 54.70 & 65.00 & 65.75 & 62.62 & 63.59 & 65.79 & 65.82 & \textbf{79.06} & 63.93 \\
\hdashline
\rowcolor{tablepurple}
\textbf{RubricArmor (Ours)} & \textbf{74.03} & 67.11 & \textbf{83.33} & \textbf{72.60} & \textbf{68.48} & \textbf{86.23} & \textbf{71.71} & \textbf{91.58} & 78.73 & \textbf{77.09} \\
\bottomrule
\end{tabular}%
}
\par\vspace{2pt}
\parbox{\textwidth}{\scriptsize\textit{Note:} Higher is better. Macro Avg. is the unweighted mean across the nine evaluation sets. Best results in each column are \textbf{bolded}.}
\end{table}

\subsection{Experimental Setup}
\label{sec:experimental_setup}

\textbf{Datasets and Metrics.} We evaluate on nine evaluation sets from three benchmark suites: Knowledge, Reasoning, Math, and Coding from JudgeBench~\citep{tan2025judgebench}; Chat, Math, Code, and Safety from RM-Bench~\citep{liu2025rmbench}; and Chat-Hard from RewardBench~\citep{lambert2024rewardbench}. Each instance consists of a query, a chosen response, and a rejected response. For each query, the verifier scores both responses against the generated rubric with Equation~\ref{eq:rubric_reward}, and a pair is counted as correct when the chosen response receives the higher reward, with ties receiving half credit. We report the preference accuracy on each evaluation set and the macro average across the nine sets. Since a high-quality rubric should assign a higher reward to the better response, preference accuracy serves as our measure of rubric quality.

\textbf{Baselines.} We compare RubricArmor against two groups of baselines: (1) \textbf{LLM-based rubric generation} methods, including TICK~\citep{cook2024ticking} and RocketEval~\citep{wei2025rocketeval}, which translate query requirements into evaluation checklists, and ChasingTail~\citep{zhang2026chasing} and RRD-Uniform~\citep{shen2026rethinking}, which progressively refine the generated criteria; and (2) \textbf{Trained rubric generators}, including Rubric-ARM~\citep{xu2026alternating}, Rubric-RM~\citep{liu2025openrubrics}, and Rubric-ARROW~\citep{jiang2026rubricarrow}, which fine-tune 8B LLMs to generate rubrics. We also report Direct, which prompts an LLM to generate a rubric directly from the query.

\textbf{Implementation Details.} Unless otherwise specified, Gemini-3.1-Flash-Lite serves as the rubric generator backbone, i.e., the LLM behind all agents of RubricArmor and all LLM-based rubric generation methods. For fair comparison, GPT-4.1-mini serves as the verifier for all methods with the same criterion-wise judgments and reward aggregation. We set the maximum number of evolution rounds to $T=7$ and the patience to $P=3$. More details are provided in Appendix~\ref{app:reproduction}.

\subsection{Main Results (RQ1)}
\label{sec:overall_performance}

In this section, we compare RubricArmor with competitive baselines on nine evaluation sets. The results are presented in Table~\ref{tab:main_results} with key observations as follows.
\ding{182} \textbf{Direct exhibits limited performance.} Direct yields the lowest macro accuracy among LLM-based rubric generation methods and performs close to random guessing on JudgeBench Reasoning. The limited performance indicates that rubrics generated directly from the query often omit essential criteria or leave criteria underspecified (Section~\ref{sec:motivation}) and thus produce unreliable rewards.
\ding{183} \textbf{LLM-based rubric generation methods exhibit improved rubric quality.} All four methods surpass Direct in macro accuracy through finer-grained checklists or progressive refinement. However, each method still underperforms Direct on some evaluation sets, as the criteria are constructed or refined without identifying the response defects that the rubric fails to detect.
\ding{184} \textbf{Trained rubric generators do not transfer across domains.} Trained rubric generators perform competitively on Chat-Hard but fall below Direct in macro accuracy, especially on RM-Bench Math and Safety. The uneven performance indicates that the rubric generation ability acquired through fine-tuning is tied to the training domains.
\ding{185} \textbf{RubricArmor achieves the best and most consistent performance.} RubricArmor attains the highest macro accuracy and outperforms Direct on all nine evaluation sets. In particular, RubricArmor exceeds the strongest baseline ChasingTail by 3.12 points in macro accuracy. These results demonstrate that RubricArmor substantially and consistently improves the quality of LLM-based rubric generation.

\begin{table}[t]
\centering
\caption{Ablations of response construction and the adversarial objective.}
\label{tab:ablation}
\scriptsize
\setlength{\tabcolsep}{2.1pt}
\renewcommand{\arraystretch}{1.12}
\resizebox{\textwidth}{!}{%
\begin{tabular}{l*{12}{c}}
\toprule
\multirow{2}{*}{\textbf{Method}}
& \multirow{2}{*}{\textbf{Resp.}}
& \multirow{2}{*}{\textbf{Adv.}}
& \multicolumn{4}{c}{\textbf{JudgeBench}}
& \multicolumn{4}{c}{\textbf{RM-Bench}}
& \textbf{RewardBench} & \textbf{Macro} \\
\cmidrule(lr){4-7}\cmidrule(lr){8-11}\cmidrule(lr){12-12}\cmidrule(lr){13-13}
&&& Knowledge & Reasoning & Math & Coding
& Chat & Math & Code & Safety & Chat-Hard & Avg. \\
\midrule
Direct & \XSolidBrush & \XSolidBrush & 62.01 & 53.69 & 70.56 & 69.18 & 64.38 & 77.59 & 69.03 & 89.91 & 74.67 & 70.11 \\
RubricArmor$^{\dagger}$ & \XSolidBrush & \Checkmark & 68.99 & 61.41 & 72.22 & 69.86 & 64.60 & 80.01 & 70.81 & 88.23 & 69.08 & 71.69 \\
RubricArmor$^{\ddagger}$ & \Checkmark & \XSolidBrush & 70.45 & 62.75 & 73.33 & 71.23 & 64.81 & 83.23 & 69.01 & 89.51 & 70.83 & 72.80 \\
\hdashline
\rowcolor{tablepurple}
\textbf{RubricArmor (Full)} & \Checkmark & \Checkmark & \textbf{74.03} & \textbf{67.11} & \textbf{83.33} & \textbf{72.60} & \textbf{68.48} & \textbf{86.23} & \textbf{71.71} & \textbf{91.58} & \textbf{78.73} & \textbf{77.09} \\
\bottomrule
\end{tabular}%
}
\par\vspace{2pt}
\parbox{\textwidth}{\scriptsize\textit{Note:} Resp.\ denotes response construction; Adv.\ denotes the adversarial objective. \Checkmark: enabled; \XSolidBrush: disabled. Macro Avg. averages the nine evaluation sets. Best results in each column are \textbf{bolded}.}
\end{table}

\begin{figure}[t]
\centering
\includegraphics[width=\linewidth]{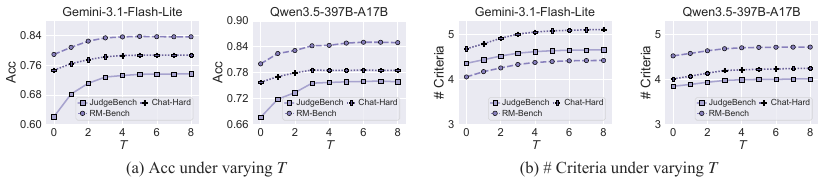}
\caption{Performance under varying maximum number of evolution rounds $T$ with two rubric generator backbones. For JudgeBench and RM-Bench, Acc denotes the micro accuracy, which weights each evaluation set by its number of comparisons.}
\label{fig:evolution_sensitivity}
\end{figure}

\subsection{Ablation Study (RQ2)}
\label{sec:ablation}

In this section, we dissect the two designs of Attack Discovery: constructing a concrete response (response construction) that is deliberately low-quality yet satisfies the current rubric (adversarial objective). The results are presented in Table~\ref{tab:ablation} with key observations as follows.
\ding{182} \textbf{Constructing concrete responses grounds rubric repair.} Specifically, in RubricArmor$^{\dagger}$, the attacker only describes potential defects without constructing a response, which amounts to asking an LLM to critique its own rubric, with the described defects directly used for repair. The macro accuracy drops to 71.69\% and even falls below Direct on Chat-Hard, suggesting that repairs based on unverified defects can penalize valid responses.
\ding{183} \textbf{The adversarial objective supplies the key evidence for repair.} Specifically, in RubricArmor$^{\ddagger}$, the attacker simply constructs a response that satisfies the current rubric without deliberate defects, and the response is used for repair without confirmation by the quality judge. The macro accuracy drops to 72.80\%, comparable to LLM-based rubric generation methods, indicating that such revision adds little beyond existing refinement.
\ding{184} \textbf{The two designs are complementary.} The gain of RubricArmor over Direct exceeds the sum of the gains of both variants, since the adversarial objective decides which defects to seek and response construction makes each defect verifiable.

\subsection{Generalization across Rubric Generator Backbones (RQ3)}
\label{sec:generalization}

\noindent
\begin{minipage}[t]{0.47\textwidth}
\vspace{0pt}
In this section, we evaluate the generalizability of RubricArmor across rubric generator backbones of different families and sizes, specifically Gemini-3.1-Flash-Lite, GPT-4.1, DeepSeek-v4-Flash, Qwen3.5-397B-A17B, and GLM-5.1. For each backbone, Direct remains the initial rubric of RubricArmor, and GPT-4.1-mini remains the verifier. The results are presented in Table~\ref{tab:generalization} with key observations as follows.
\ding{182} \textbf{RubricArmor generalizes across backbone families.} RubricArmor improves over Direct on all three benchmark suites under every backbone, indicating that the gains stem from adversarial evolution rather than from a particular backbone family.
\ding{183} \textbf{RubricArmor generalizes across backbone capabilities.} RubricArmor consistently improves over Direct under both weaker and stronger backbones. This shows that the gains complement rather than depend on backbone capacity.
\end{minipage}\hfill
\begin{minipage}[t]{0.50\textwidth}
\vspace{0pt}
\centering
\setlength{\abovecaptionskip}{0pt}
\captionof{table}{Generalization across rubric generator backbones.}
\label{tab:generalization}
\scriptsize
\setlength{\tabcolsep}{0.9pt}
\renewcommand{\arraystretch}{1.08}
\begin{tabularx}{\linewidth}{l*{6}{>{\centering\arraybackslash}X}}
\toprule
& \multicolumn{2}{c}{\textbf{JudgeBench}} & \multicolumn{2}{c}{\textbf{RM-Bench}} & \multicolumn{2}{c}{\textbf{Chat-Hard}} \\
& \multicolumn{2}{c}{Micro} & \multicolumn{2}{c}{Micro} & \multicolumn{2}{c}{Accuracy} \\
\cmidrule(lr){2-3}\cmidrule(lr){4-5}\cmidrule(lr){6-7}
\textbf{Method} & Value & $\Delta$ & Value & $\Delta$ & Value & $\Delta$ \\
\midrule
\multicolumn{7}{l}{\cellcolor{tablegray}\textbf{\textit{Backbone: Gemini-3.1-Flash-Lite}}} \\
Direct & 62.10 & --- & 78.93 & --- & 74.67 & --- \\
\rowcolor{tablepurple}
\textbf{RubricArmor} & \textbf{73.63} & $18.6\%\!\uparrow$ & \textbf{83.69} & $6.0\%\!\uparrow$ & \textbf{78.73} & $5.4\%\!\uparrow$ \\
\midrule
\multicolumn{7}{l}{\cellcolor{tablegray}\textbf{\textit{Backbone: GPT-4.1}}} \\
Direct & 61.53 & --- & 71.47 & --- & 75.66 & --- \\
\rowcolor{tablepurple}
\textbf{RubricArmor} & \textbf{64.76} & $5.2\%\!\uparrow$ & \textbf{74.38} & $4.1\%\!\uparrow$ & \textbf{77.96} & $3.0\%\!\uparrow$ \\
\midrule
\multicolumn{7}{l}{\cellcolor{tablegray}\textbf{\textit{Backbone: DeepSeek-v4-Flash}}} \\
Direct & 83.47 & --- & 83.35 & --- & 74.34 & --- \\
\rowcolor{tablepurple}
\textbf{RubricArmor} & \textbf{87.50} & $4.8\%\!\uparrow$ & \textbf{85.92} & $3.1\%\!\uparrow$ & \textbf{80.48} & $8.3\%\!\uparrow$ \\
\midrule
\multicolumn{7}{l}{\cellcolor{tablegray}\textbf{\textit{Backbone: Qwen3.5-397B-A17B}}} \\
Direct & 67.66 & --- & 80.04 & --- & 75.77 & --- \\
\rowcolor{tablepurple}
\textbf{RubricArmor} & \textbf{76.05} & $12.4\%\!\uparrow$ & \textbf{85.04} & $6.3\%\!\uparrow$ & \textbf{78.51} & $3.6\%\!\uparrow$ \\
\midrule
\multicolumn{7}{l}{\cellcolor{tablegray}\textbf{\textit{Backbone: GLM-5.1}}} \\
Direct & 85.40 & --- & 82.24 & --- & 72.59 & --- \\
\rowcolor{tablepurple}
\textbf{RubricArmor} & \textbf{86.29} & $1.0\%\!\uparrow$ & \textbf{85.26} & $3.7\%\!\uparrow$ & \textbf{77.96} & $7.4\%\!\uparrow$ \\
\bottomrule
\end{tabularx}
\par\vspace{2pt}
\parbox{\linewidth}{\scriptsize\textit{Note:} Values: accuracy (\%). Micro averages use sample-count weights; $\Delta$ is relative improvement over Direct.}
\end{minipage}
\par
\vspace{2pt}
\noindent Collectively, RubricArmor is model-agnostic, as confirmed across verifiers in Appendix~\ref{app:verifier_generalization}.

\subsection{Hyperparameter Sensitivity (RQ4)}
\label{sec:evolution_dynamics}

In this section, we examine the impact of the maximum number of evolution rounds $T$ on the performance of RubricArmor. Specifically, we vary $T$ from 0 to 8 under two rubric generator backbones, Gemini-3.1-Flash-Lite and Qwen3.5-397B-A17B. The results are presented in Figure~\ref{fig:evolution_sensitivity} with key observations as follows.
\ding{182} \textbf{The gains of RubricArmor saturate as $T$ increases.} For both backbones, the accuracy on all three benchmark suites rises rapidly in early rounds and then plateaus. This is attributed to the attack--repair loop. Each accepted revision enables the rubric to detect a confirmed defect, so fewer attacks succeed and the rubric stabilizes in later rounds. Such saturation empirically demonstrates the convergence of adversarial evolution.
\ding{183} \textbf{The gains of RubricArmor stem from rubric quality rather than more criteria.} As the accuracy rises, the average number of criteria per rubric remains nearly stable for both backbones (e.g., from 4.36 to 4.66 on JudgeBench with Gemini-3.1-Flash-Lite). This indicates that adversarial evolution improves the quality of the criteria rather than accumulating more criteria. The computational cost is reported in Appendix~\ref{app:cost}.

\begin{table}[t]
\setlength{\abovecaptionskip}{0pt}
\setlength{\belowcaptionskip}{6pt}
\caption{Downstream RL results.}
\label{tab:downstream_rl}
\centering
\renewcommand{\arraystretch}{1.00}
\fontsize{8.3}{10}\selectfont
\def\experimentrows{
\toprule
\multirow{3}{*}{\textbf{Method}} & \multicolumn{2}{c}{\textbf{BiGGen-Bench}} & \multicolumn{4}{c}{\textbf{AlpacaEval 2.0}} & \multicolumn{4}{c}{\textbf{Arena-Hard}} \\
\cmidrule(lr){2-3}\cmidrule(lr){4-7}\cmidrule(lr){8-11}
 & \multicolumn{2}{c}{Score} & \multicolumn{2}{c}{Vanilla} & \multicolumn{2}{c}{LC} & \multicolumn{2}{c}{Vanilla} & \multicolumn{2}{c}{SC} \\
\cmidrule(lr){2-3}\cmidrule(lr){4-5}\cmidrule(lr){6-7}\cmidrule(lr){8-9}\cmidrule(lr){10-11}
 & Value & $\Delta$ & Value & $\Delta$ & Value & $\Delta$ & Value & $\Delta$ & Value & $\Delta$ \\
\midrule
\rowcolor{tablegray}\multicolumn{11}{l}{\textbf{\textit{Policy Model: Qwen2.5-3B-Instruct}}} \\
\rowcolor{white}
Original & 63.18 & - & 16.48 & - & 15.62 & - & 26.11 & - & 26.55 & - \\
\rowcolor{white}
Direct & 63.41 & 0.36\%$\uparrow$ & 19.81 & 20.21\%$\uparrow$ & 18.18 & 16.39\%$\uparrow$ & 26.41 & 1.15\%$\uparrow$ & 28.10 & 5.84\%$\uparrow$ \\
\rowcolor{white}
TICK & 63.84 & 1.04\%$\uparrow$ & 18.38 & 11.53\%$\uparrow$ & 17.29 & 10.69\%$\uparrow$ & 26.69 & 2.22\%$\uparrow$ & 28.14 & 5.99\%$\uparrow$ \\
\rowcolor{white}
RocketEval & 63.27 & 0.14\%$\uparrow$ & 21.02 & 27.55\%$\uparrow$ & 18.19 & 16.45\%$\uparrow$ & 28.10 & 7.62\%$\uparrow$ & 28.83 & 8.59\%$\uparrow$ \\
\rowcolor{white}
ChasingTail & 64.39 & 1.92\%$\uparrow$ & 23.61 & 43.26\%$\uparrow$ & 22.84 & 46.22\%$\uparrow$ & 28.54 & 9.31\%$\uparrow$ & 28.31 & 6.63\%$\uparrow$ \\
\rowcolor{tablepurple}
\textbf{RubricArmor} & \textbf{65.21} & \textbf{3.21\%$\uparrow$} & \textbf{25.13} & \textbf{52.49\%$\uparrow$} & \textbf{24.34} & \textbf{55.83\%$\uparrow$} & \textbf{29.50} & \textbf{12.98\%$\uparrow$} & \textbf{29.23} & \textbf{10.09\%$\uparrow$} \\
\midrule
\rowcolor{tablegray}\multicolumn{11}{l}{\textbf{\textit{Policy Model: Qwen2.5-7B-Instruct}}} \\
\rowcolor{white}
Original & 68.11 & - & 27.70 & - & 28.93 & - & 50.84 & - & 53.52 & - \\
\rowcolor{white}
Direct & 68.83 & 1.06\%$\uparrow$ & 32.89 & 18.74\%$\uparrow$ & 30.24 & 4.53\%$\uparrow$ & 52.02 & 2.32\%$\uparrow$ & 54.11 & 1.10\%$\uparrow$ \\
\rowcolor{white}
TICK & 69.22 & 1.63\%$\uparrow$ & 28.29 & 2.13\%$\uparrow$ & 29.18 & 0.86\%$\uparrow$ & 50.67 & 0.33\%$\downarrow$ & 53.23 & 0.54\%$\downarrow$ \\
\rowcolor{white}
RocketEval & 70.75 & 3.88\%$\uparrow$ & 36.24 & 30.83\%$\uparrow$ & 32.38 & 11.93\%$\uparrow$ & 54.79 & 7.77\%$\uparrow$ & 55.22 & 3.18\%$\uparrow$ \\
\rowcolor{white}
ChasingTail & 71.62 & 5.15\%$\uparrow$ & \textbf{39.99} & \textbf{44.37\%$\uparrow$} & 31.89 & 10.23\%$\uparrow$ & \textbf{56.72} & \textbf{11.57\%$\uparrow$} & 54.99 & 2.75\%$\uparrow$ \\
\rowcolor{tablepurple}
\textbf{RubricArmor} & \textbf{72.05} & \textbf{5.78\%$\uparrow$} & 39.64 & 43.10\%$\uparrow$ & \textbf{33.64} & \textbf{16.28\%$\uparrow$} & 56.66 & 11.45\%$\uparrow$ & \textbf{56.60} & \textbf{5.75\%$\uparrow$} \\
\midrule
\rowcolor{tablegray}\multicolumn{11}{l}{\textbf{\textit{Policy Model: Qwen3-4B}}} \\
\rowcolor{white}
Original & 76.49 & - & 48.84 & - & 39.95 & - & 73.08 & - & 72.01 & - \\
\rowcolor{white}
Direct & 77.26 & 1.01\%$\uparrow$ & 51.94 & 6.35\%$\uparrow$ & 41.73 & 4.46\%$\uparrow$ & 73.04 & 0.05\%$\downarrow$ & 72.45 & 0.61\%$\uparrow$ \\
\rowcolor{white}
TICK & 77.77 & 1.67\%$\uparrow$ & 50.73 & 3.87\%$\uparrow$ & 42.15 & 5.51\%$\uparrow$ & 72.55 & 0.73\%$\downarrow$ & 71.20 & 1.12\%$\downarrow$ \\
\rowcolor{white}
RocketEval & 77.26 & 1.01\%$\uparrow$ & 53.81 & 10.18\%$\uparrow$ & 40.53 & 1.45\%$\uparrow$ & 75.82 & 3.75\%$\uparrow$ & 74.03 & 2.81\%$\uparrow$ \\
\rowcolor{white}
ChasingTail & 78.03 & 2.01\%$\uparrow$ & 52.76 & 8.03\%$\uparrow$ & 41.90 & 4.88\%$\uparrow$ & 75.45 & 3.24\%$\uparrow$ & 73.46 & 2.01\%$\uparrow$ \\
\rowcolor{tablepurple}
\textbf{RubricArmor} & \textbf{78.67} & \textbf{2.85\%$\uparrow$} & \textbf{53.91} & \textbf{10.38\%$\uparrow$} & \textbf{42.43} & \textbf{6.21\%$\uparrow$} & \textbf{76.17} & \textbf{4.23\%$\uparrow$} & \textbf{75.62} & \textbf{5.01\%$\uparrow$} \\
\bottomrule
}
\setlength{\tabcolsep}{0pt}
\setbox0=\hbox{\begin{tabular}{l*{10}{c}}\experimentrows\end{tabular}}
\dimen0=\dimexpr\linewidth-\wd0\relax
\divide\dimen0 by 22
\setlength{\tabcolsep}{\dimen0}
\begin{tabular}{l*{10}{c}}\experimentrows\end{tabular}
\par\vspace{3pt}\raggedright\footnotesize
\textit{Note:} LC and SC denote length-controlled and style-controlled win rates. $\Delta$ indicates the relative change over the Original policy in percentage. The best performances are \textbf{bolded}.
\end{table}

\subsection{Downstream Rubric-Based RL (RQ5)}
\label{sec:downstream_rl}

In this section, we investigate the downstream RL performance of RubricArmor. We select representative LLM-based rubric generation baselines from Table~\ref{tab:main_results}, generate rubrics for 2K queries from WildChat~\citep{zhao2024wildchat}, and fine-tune three policy models via GRPO~\citep{shao2024deepseekmath} with the generated rubrics as rewards. We evaluate the policies on BiGGen-Bench~\citep{kim2025biggen}, AlpacaEval 2.0~\citep{li2023alpacaeval,dubois2024length}, and Arena-Hard~\citep{li2024crowdsourced} with GPT-4.1 as the judge. The results are presented in Table~\ref{tab:downstream_rl} with key observations as follows.
\ding{182} \textbf{Rubric-based RL improves the original policies.} All methods improve the BiGGen-Bench scores and AlpacaEval 2.0 win rates of the three policy models over the original policies, confirming that rubric rewards translate into better policy behavior after GRPO.
\ding{183} \textbf{RubricArmor achieves the best downstream RL performance across policy models.} RubricArmor leads on 13 of the 15 policy--metric combinations. On Qwen2.5-3B-Instruct, RubricArmor improves the AlpacaEval 2.0 length-controlled (LC) win rate by 55.83\% over the original policy and by a further 6.57\% over the strongest baseline ChasingTail. These gains highlight that the improved rubric quality of RubricArmor translates into better downstream performance, as the higher-quality rubrics provide more reliable rewards for policy optimization.
\ding{184} \textbf{The gains of RubricArmor hold under length and style control.} RubricArmor leads on all length-controlled and style-controlled win rates. On Qwen2.5-7B-Instruct, ChasingTail leads on vanilla win rates but falls behind RubricArmor once length and style are controlled. This indicates that policies trained with RubricArmor rubrics improve by properly completing the task rather than by producing longer or more heavily formatted responses. Since inflating length and formatting is a common form of reward hacking~\citep{mahmoud2026hacking}, this result provides indirect evidence that RubricArmor mitigates reward hacking in rubric-based RL.

\section{Conclusion}
\label{sec:conclusion}

In this paper, we investigated the vulnerability of rubrics directly generated by LLMs to reward hacking, where omitted or underspecified criteria allow low-quality responses to obtain high rubric rewards. To address this limitation, we proposed RubricArmor, an adversarial framework that exposes and mitigates potential reward hacking before RL through adversarial evolution, which alternates an attack step that constructs adversarial responses with a repair step that revises the rubric to detect the exposed defects. Extensive experiments confirm that RubricArmor exceeds the strongest baseline by 3.12 points in macro accuracy on nine preference evaluation sets and translates into the best downstream RL performance in 13 of 15 settings across three policy models.

\textbf{Limitations \& Future Work.} Two limitations of this work warrant future investigation: (1) RubricArmor requires multiple rounds of LLM calls to construct each rubric. Although the cost is incurred once before policy optimization, reducing the cost remains a valuable direction, for example by sharing confirmed defects across similar queries. (2) RubricArmor relies on LLM agents to confirm defects, so errors of the quality judge can propagate into the revised rubric, motivating stronger verification mechanisms.

\FloatBarrier
\bibliography{iclr2027_conference}
\bibliographystyle{arxiv_style/icml2026}

\clearpage
\appendix

\section{Related Work}
\label{app:related_work}

This work studies LLM-based rubric generation for rubric-based RL, focusing on the reward hacking caused by rubric defects. The related literature broadly falls into three lines: (1) rubric-based RL, which uses rubrics as rewards for policy optimization; (2) LLM-based rubric generation, which constructs rubrics by prompting LLMs without training a rubric generator; and (3) reward hacking, which studies how policies exploit imperfect rewards and how to mitigate the exploitation. RubricArmor differs from these lines in exposing and mitigating potential reward hacking at the rubric generation stage before RL. Specifically, RubricArmor simulates the reward hacking of the policy and repairs the rubric to detect the exposed response defects.

\subsection{Rubric-Based Reinforcement Learning}
Rubric-based RL provides interpretable and task-specific rewards by evaluating responses against explicit criteria tailored to each query~\citep{gunjal2025rubrics,he2025rifl}. A prevalent strategy constructs \textbf{static rubrics} before training and uses the rubrics as fixed rewards throughout policy optimization, either by synthesizing large-scale rubric datasets~\citep{li2026rubrichub,li2026ares} or by training rubric reward models that generate criteria and judge responses~\citep{liu2025openrubrics,xu2026alternating,jiang2026rubricarrow}. Since the evolving policy can exploit a fixed rubric, a more recent line of work adopts \textbf{dynamic rubrics} that are updated during RL, either by eliciting new criteria from pairwise comparisons between policy and reference responses~\citep{rezaei2025online} or by co-evolving the policy and a rubric generator within each training step~\citep{ding2026evorubrics}. However, two limitations remain in rubric-based RL: (1) static rubrics are constructed without proactively guarding against reward hacking; and (2) dynamic rubrics adapt only to the behaviors that the policy has already produced during training, which couples rubric construction with the RL loop. RubricArmor addresses both limitations by exposing and repairing the response defects that a rubric fails to detect before RL, which produces a rubric that is reused throughout policy optimization and complements dynamic rubric methods.

\subsection{LLM-Based Rubric Generation}
LLM-based rubric generation constructs query-specific rubrics by prompting LLMs, requiring neither expert annotation nor a trained rubric generator. One line of work performs \textbf{checklist generation}, which translates query requirements into binary evaluation questions for an LLM judge~\citep{cook2024ticking,lee2025checkeval,wei2025rocketeval}. Another line of work performs \textbf{progressive refinement}, which improves the granularity and coverage of the generated criteria by distinguishing among diverse high-quality responses~\citep{zhang2026chasing}, decomposing coarse criteria into fine-grained ones~\citep{shen2026rethinking,li2026rubrichub}, expanding each question into scenarios and perspectives~\citep{gao2026qworld}, or eliciting criteria from multiple complementary roles~\citep{fu2026many}. Both lines of work improve how comprehensively and precisely a rubric describes an ideal response. However, both lines construct or refine the criteria without adversarial responses that exploit the rubric, and thus do not proactively guard against reward hacking. RubricArmor instead treats each generated rubric as the target of attack: the attack step constructs adversarial responses that simulate the reward hacking of the policy, and the repair step revises the rubric to detect the confirmed defects.

\subsection{Reward Hacking}
\textbf{Reward hacking} occurs when increasing an imperfect proxy reward decreases the true objective~\citep{skalse2022defining}. In RLHF, optimizing a learned reward model beyond a certain point decreases the reward measured by a gold reward model~\citep{gao2023scaling}. To mitigate such overoptimization, prevalent methods regularize the reward model, for example through reward model ensembles~\citep{coste2024reward} or information-theoretic reward modeling~\citep{miao2024inform}. Rubric-based rewards are likewise exploitable, as policies learn to satisfy the letter of a rubric while degrading holistic quality~\citep{mahmoud2026hacking} and exploit latent biases of the LLM judge~\citep{wang2026reproducing}. Recent benchmarks further stress-test generated rubrics with adversarial answers on impossible tasks~\citep{qin2026impossiblerubrics}. To mitigate such exploitation during training, Rubric Dropout randomly drops a subset of criteria at every step so that the policy never optimizes the same rubric twice~\citep{yang2026rubricdropout}. Existing studies diagnose reward hacking, and existing mitigations act during policy optimization, leaving the rubric defects that enable reward hacking unrepaired. RubricArmor intervenes at an earlier stage by repairing the rubric defects before RL, and is thus complementary to training-time mitigation.

\section{More Experimental Results}
\label{app:more_results}

In this section, we present supplementary results deferred from the main text due to space constraints. First, we extend the generalization study in Section~\ref{sec:generalization} to verifier backbones (Appendix~\ref{app:verifier_generalization}). Second, we report the computational cost of RubricArmor across evolution rounds (Appendix~\ref{app:cost}). Third, we evaluate the out-of-domain generalization of the RL-trained policies on HealthBench-Hard (Appendix~\ref{app:ood_generalization}). 


\subsection{Generalization across Verifier Backbones}
\label{app:verifier_generalization}

\noindent
\begin{minipage}[t]{0.47\textwidth}
\vspace{0pt}
In this subsection, we evaluate the generalizability of RubricArmor across verifier backbones of different families and capabilities. For each verifier backbone, Direct and RubricArmor reuse the rubrics in Table~\ref{tab:main_results}. The results are presented in Table~\ref{tab:verifier_generalization} with key observations as follows.
\ding{182} \textbf{RubricArmor generalizes across verifier families.} RubricArmor improves over Direct on all three benchmark suites under every verifier backbone, indicating that RubricArmor improves the rubrics themselves rather than fitting a particular verifier family.
\ding{183}~\textbf{RubricArmor generalizes across verifier capabilities.} RubricArmor improves over Direct under both weaker and stronger verifiers, with the largest relative gains under the weakest verifier GPT-4o-mini on JudgeBench and RM-Bench. This suggests that self-contained criteria reduce the knowledge that a verifier must supply.
\end{minipage}\hfill
\begin{minipage}[t]{0.50\textwidth}
\vspace{0pt}
\centering
\setlength{\abovecaptionskip}{0pt}
\captionof{table}{Generalization across verifier backbones.}
\label{tab:verifier_generalization}
\scriptsize
\setlength{\tabcolsep}{0.9pt}
\renewcommand{\arraystretch}{1.08}
\begin{tabularx}{\linewidth}{l*{6}{>{\centering\arraybackslash}X}}
\toprule
& \multicolumn{2}{c}{\textbf{JudgeBench}} & \multicolumn{2}{c}{\textbf{RM-Bench}} & \multicolumn{2}{c}{\textbf{Chat-Hard}} \\
& \multicolumn{2}{c}{Micro} & \multicolumn{2}{c}{Micro} & \multicolumn{2}{c}{Accuracy} \\
\cmidrule(lr){2-3}\cmidrule(lr){4-5}\cmidrule(lr){6-7}
\textbf{Method} & Value & $\Delta$ & Value & $\Delta$ & Value & $\Delta$ \\
\midrule
\multicolumn{7}{l}{\cellcolor{tablegray}\textbf{\textit{Backbone: GPT-4.1}}} \\
Direct & 65.56 & --- & 75.92 & --- & 74.89 & --- \\
\rowcolor{tablepurple}
\textbf{RubricArmor} & \textbf{73.06} & $11.4\%\!\uparrow$ & \textbf{79.98} & $5.3\%\!\uparrow$ & \textbf{77.74} & $3.8\%\!\uparrow$ \\
\midrule
\multicolumn{7}{l}{\cellcolor{tablegray}\textbf{\textit{Backbone: Gemini-3.5-Flash}}} \\
Direct & 73.23 & --- & 79.81 & --- & 75.33 & --- \\
\rowcolor{tablepurple}
\textbf{RubricArmor} & \textbf{79.92} & $9.1\%\!\uparrow$ & \textbf{83.82} & $5.0\%\!\uparrow$ & \textbf{78.84} & $4.7\%\!\uparrow$ \\
\midrule
\multicolumn{7}{l}{\cellcolor{tablegray}\textbf{\textit{Backbone: GPT-4o-mini}}} \\
Direct & 63.39 & --- & 73.63 & --- & 74.01 & --- \\
\rowcolor{tablepurple}
\textbf{RubricArmor} & \textbf{71.45} & $12.7\%\!\uparrow$ & \textbf{78.59} & $6.7\%\!\uparrow$ & \textbf{74.67} & $0.9\%\!\uparrow$ \\
\midrule
\multicolumn{7}{l}{\cellcolor{tablegray}\textbf{\textit{Backbone: GLM-5.1}}} \\
Direct & 71.53 & --- & 78.73 & --- & 75.88 & --- \\
\rowcolor{tablepurple}
\textbf{RubricArmor} & \textbf{79.03} & $10.5\%\!\uparrow$ & \textbf{83.36} & $5.9\%\!\uparrow$ & \textbf{78.07} & $2.9\%\!\uparrow$ \\
\midrule
\multicolumn{7}{l}{\cellcolor{tablegray}\textbf{\textit{Backbone: DeepSeek-v4-Flash}}} \\
Direct & 65.40 & --- & 76.78 & --- & 74.01 & --- \\
\rowcolor{tablepurple}
\textbf{RubricArmor} & \textbf{72.58} & $11.0\%\!\uparrow$ & \textbf{80.53} & $4.9\%\!\uparrow$ & \textbf{77.41} & $4.6\%\!\uparrow$ \\
\bottomrule
\end{tabularx}
\par\vspace{2pt}
\parbox{\linewidth}{\scriptsize\textit{Note:} Values: accuracy (\%). Micro averages use sample-count weights; $\Delta$ is relative improvement over Direct.}
\end{minipage}
\par

\subsection{Computational Cost}
\label{app:cost}

\noindent
\begin{minipage}[t]{0.49\textwidth}
\vspace{0pt}
In this subsection, we measure the computational cost of RubricArmor by the total tokens and elapsed time per query, and the results are presented in Figure~\ref{fig:evolution_cost}. \textbf{Early stopping bounds the cost of adversarial evolution.} The cost grows quickly in the first three rounds and flattens afterwards, since more than half of the queries stop after three consecutive rounds without an accepted revision. Consequently, increasing the maximum number of evolution rounds $T$ adds cost only for the few queries whose rubrics are still being revised in later rounds.
\end{minipage}\hfill
\begin{minipage}[t]{0.49\textwidth}
\vspace{0pt}
\centering
\includegraphics[width=\linewidth]{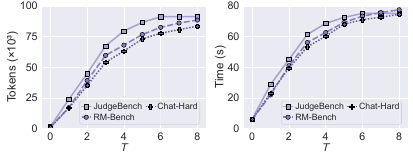}
\captionof{figure}{Cumulative total tokens and elapsed time per query under varying maximum number of evolution rounds $T$.}
\label{fig:evolution_cost}
\end{minipage}
\par

\subsection{Out-of-Domain Policy Generalization}
\label{app:ood_generalization}

In this subsection, we evaluate whether the policy improvements from RubricArmor generalize beyond the training domain. We evaluate the WildChat-trained policies in Section~\ref{sec:downstream_rl} on HealthBench-Hard~\citep{arora2025healthbench}, which comprises 1,000 healthcare conversations with physician-written criteria, using GPT-4.1 as the judge. To better distinguish methods whose aggregate scores would otherwise be clipped to zero, we clip each sample--axis score to $[0,1]$ before averaging. The results are presented in Table~\ref{tab:healthbench} with key observations as follows.
\ding{182} \textbf{RubricArmor improves policies beyond the training domain.} RubricArmor improves over the original policies on all five axes for all three policy models, whereas each of the other methods underperforms the original policy on at least one policy--axis combination. This indicates that the improved rubric quality of RubricArmor yields policy gains that transfer from general-purpose tasks to a specialized domain.
\ding{183}~\textbf{RubricArmor remains the most effective rubric reward under domain shift.} RubricArmor achieves the highest average score for all three policy models and the best score on 9 of the 15 policy--axis combinations, including Communication Quality for every policy model. Since the physician-written criteria are never used in rubric generation or policy optimization, these gains are unlikely to stem from exploiting the training rubrics, which provides further indirect evidence that RubricArmor mitigates reward hacking in rubric-based RL.

\begin{table}[!ht]
\setlength{\abovecaptionskip}{0pt}
\setlength{\belowcaptionskip}{6pt}
\caption{Out-of-domain generalization of RL-trained policies on HealthBench-Hard.}
\label{tab:healthbench}
\centering
\renewcommand{\arraystretch}{1.00}
\fontsize{7.8}{10}\selectfont
\def\experimentrows{
\toprule
\multirow{2}{*}{\textbf{Method}} & \multicolumn{2}{c}{\textbf{Accuracy}} & \multicolumn{2}{c}{\shortstack{\textbf{Communication}\\\textbf{Quality}}} & \multicolumn{2}{c}{\textbf{Completeness}} & \multicolumn{2}{c}{\shortstack{\textbf{Context}\\\textbf{Awareness}}} & \multicolumn{2}{c}{\shortstack{\textbf{Instruction}\\\textbf{Following}}} & \multicolumn{2}{c}{\textbf{Average}} \\
\cmidrule(lr){2-3}\cmidrule(lr){4-5}\cmidrule(lr){6-7}\cmidrule(lr){8-9}\cmidrule(lr){10-11}\cmidrule(lr){12-13}
 & Value & $\Delta$ & Value & $\Delta$ & Value & $\Delta$ & Value & $\Delta$ & Value & $\Delta$ & Value & $\Delta$ \\
\midrule
\rowcolor{tablegray}\multicolumn{13}{l}{\textbf{\textit{Policy Model: Qwen2.5-3B-Instruct}}} \\
\rowcolor{white}
Original & 19.66 & - & 46.64 & - & 11.12 & - & 13.69 & - & 34.84 & - & 25.19 & - \\
\rowcolor{white}
Direct & \textbf{20.43} & \textbf{3.92\%$\uparrow$} & 45.23 & 3.02\%$\downarrow$ & 12.36 & 11.15\%$\uparrow$ & 13.45 & 1.75\%$\downarrow$ & 35.56 & 2.07\%$\uparrow$ & 25.41 & 0.86\%$\uparrow$ \\
\rowcolor{white}
TICK & 19.85 & 0.97\%$\uparrow$ & 46.27 & 0.79\%$\downarrow$ & 11.14 & 0.18\%$\uparrow$ & 14.24 & 4.02\%$\uparrow$ & 35.40 & 1.61\%$\uparrow$ & 25.38 & 0.75\%$\uparrow$ \\
\rowcolor{white}
RocketEval & 20.22 & 2.85\%$\uparrow$ & 48.00 & 2.92\%$\uparrow$ & 12.83 & 15.38\%$\uparrow$ & 13.89 & 1.46\%$\uparrow$ & 38.06 & 9.24\%$\uparrow$ & 26.60 & 5.60\%$\uparrow$ \\
\rowcolor{white}
ChasingTail & 19.73 & 0.36\%$\uparrow$ & 45.73 & 1.95\%$\downarrow$ & 12.65 & 13.76\%$\uparrow$ & 14.13 & 3.21\%$\uparrow$ & \textbf{38.34} & \textbf{10.05\%$\uparrow$} & 26.12 & 3.68\%$\uparrow$ \\
\rowcolor{tablepurple}
\textbf{RubricArmor} & 20.03 & 1.88\%$\uparrow$ & \textbf{49.12} & \textbf{5.32\%$\uparrow$} & \textbf{13.79} & \textbf{24.01\%$\uparrow$} & \textbf{14.99} & \textbf{9.50\%$\uparrow$} & 38.03 & 9.16\%$\uparrow$ & \textbf{27.19} & \textbf{7.95\%$\uparrow$} \\
\midrule
\rowcolor{tablegray}\multicolumn{13}{l}{\textbf{\textit{Policy Model: Qwen2.5-7B-Instruct}}} \\
\rowcolor{white}
Original & 21.36 & - & 52.70 & - & 13.61 & - & 15.27 & - & 41.60 & - & 28.91 & - \\
\rowcolor{white}
Direct & 22.30 & 4.40\%$\uparrow$ & 53.21 & 0.97\%$\uparrow$ & 15.22 & 11.83\%$\uparrow$ & 15.91 & 4.19\%$\uparrow$ & 41.37 & 0.55\%$\downarrow$ & 29.60 & 2.40\%$\uparrow$ \\
\rowcolor{white}
TICK & 21.87 & 2.39\%$\uparrow$ & 52.23 & 0.89\%$\downarrow$ & 14.59 & 7.20\%$\uparrow$ & 15.37 & 0.65\%$\uparrow$ & 42.77 & 2.81\%$\uparrow$ & 29.37 & 1.58\%$\uparrow$ \\
\rowcolor{white}
RocketEval & \textbf{24.75} & \textbf{15.87\%$\uparrow$} & 52.34 & 0.68\%$\downarrow$ & 16.81 & 23.51\%$\uparrow$ & 17.51 & 14.67\%$\uparrow$ & 42.32 & 1.73\%$\uparrow$ & 30.75 & 6.36\%$\uparrow$ \\
\rowcolor{white}
ChasingTail & 24.42 & 14.33\%$\uparrow$ & 51.21 & 2.83\%$\downarrow$ & 16.13 & 18.52\%$\uparrow$ & 16.12 & 5.57\%$\uparrow$ & 42.16 & 1.35\%$\uparrow$ & 30.01 & 3.81\%$\uparrow$ \\
\rowcolor{tablepurple}
\textbf{RubricArmor} & 23.80 & 11.42\%$\uparrow$ & \textbf{56.34} & \textbf{6.91\%$\uparrow$} & \textbf{18.39} & \textbf{35.12\%$\uparrow$} & \textbf{18.08} & \textbf{18.40\%$\uparrow$} & \textbf{44.39} & \textbf{6.71\%$\uparrow$} & \textbf{32.20} & \textbf{11.39\%$\uparrow$} \\
\midrule
\rowcolor{tablegray}\multicolumn{13}{l}{\textbf{\textit{Policy Model: Qwen3-4B}}} \\
\rowcolor{white}
Original & 31.77 & - & 55.06 & - & 23.25 & - & 19.51 & - & 46.83 & - & 35.28 & - \\
\rowcolor{white}
Direct & 32.31 & 1.70\%$\uparrow$ & 55.23 & 0.31\%$\uparrow$ & 25.29 & 8.77\%$\uparrow$ & 19.70 & 0.97\%$\uparrow$ & 47.56 & 1.56\%$\uparrow$ & 36.02 & 2.08\%$\uparrow$ \\
\rowcolor{white}
TICK & 33.72 & 6.14\%$\uparrow$ & 56.86 & 3.27\%$\uparrow$ & 25.64 & 10.28\%$\uparrow$ & 20.88 & 7.02\%$\uparrow$ & 48.74 & 4.08\%$\uparrow$ & 37.17 & 5.34\%$\uparrow$ \\
\rowcolor{white}
RocketEval & 33.85 & 6.55\%$\uparrow$ & 54.52 & 0.98\%$\downarrow$ & \textbf{27.47} & \textbf{18.15\%$\uparrow$} & 21.06 & 7.94\%$\uparrow$ & 47.82 & 2.11\%$\uparrow$ & 36.94 & 4.70\%$\uparrow$ \\
\rowcolor{white}
ChasingTail & 32.21 & 1.38\%$\uparrow$ & 54.75 & 0.56\%$\downarrow$ & 26.76 & 15.10\%$\uparrow$ & \textbf{21.70} & \textbf{11.23\%$\uparrow$} & \textbf{48.92} & \textbf{4.46\%$\uparrow$} & 36.87 & 4.49\%$\uparrow$ \\
\rowcolor{tablepurple}
\textbf{RubricArmor} & \textbf{34.99} & \textbf{10.14\%$\uparrow$} & \textbf{58.23} & \textbf{5.76\%$\uparrow$} & 25.93 & 11.53\%$\uparrow$ & 21.63 & 10.87\%$\uparrow$ & 48.90 & 4.42\%$\uparrow$ & \textbf{37.94} & \textbf{7.52\%$\uparrow$} \\
\bottomrule
}
\setlength{\tabcolsep}{0pt}
\setbox0=\hbox{\begin{tabular}{l*{12}{c}}\experimentrows\end{tabular}}
\dimen0=\dimexpr\linewidth-\wd0\relax
\divide\dimen0 by 26
\setlength{\tabcolsep}{\dimen0}
\begin{tabular}{l*{12}{c}}\experimentrows\end{tabular}
\par\vspace{3pt}\raggedright\footnotesize
\textit{Note:} Average denotes the mean of the five axes. $\Delta$ indicates the relative change over the Original policy in percentage. The best performances are \textbf{bolded}.
\end{table}

\definecolor{caseorange}{RGB}{237,125,49}
\definecolor{casepassgreen}{RGB}{25,120,65}
\definecolor{casefailred}{RGB}{185,40,45}
\definecolor{casepromptblue}{RGB}{68,114,196}
\newcommand{\casekey}[1]{\textcolor{caseorange}{\textbf{#1}}}
\newcommand{\casepass}{\textcolor{casepassgreen}{[\ding{51}]}}
\newcommand{\casefail}{\textcolor{casefailred}{[x]}}

\section{Reproduction Details}
\label{app:reproduction}

\subsection{Dataset Descriptions}
\label{app:dataset}

In this section, we document the datasets used in our experiments, which serve two complementary purposes: (1) preference benchmarks that evaluate the quality of generated rubrics, and (2) training queries and benchmarks for downstream rubric-based RL.

\paragraph{Preference Benchmarks for Rubric Evaluation.}
We use nine evaluation sets from three benchmark suites, where each instance consists of a query, a chosen response, and a rejected response.
{\setlength{\leftmargini}{1.2em}
\begin{itemize}
    \item \textbf{JudgeBench}~\citep{tan2025judgebench}\footnote{\url{https://huggingface.co/datasets/ScalerLab/JudgeBench}}: A benchmark of challenging response pairs whose preferences are determined by objective correctness. We use the Knowledge, Reasoning, Math, and Coding sets.
    \item \textbf{RM-Bench}~\citep{liu2025rmbench}\footnote{\url{https://huggingface.co/datasets/THU-KEG/RM-Bench}}: A benchmark that evaluates sensitivity to subtle content differences and robustness to response style. Each query provides chosen and rejected responses in three styles, yielding nine comparisons per query. We use the Chat, Math, Code, and Safety sets.
    \item \textbf{RewardBench}~\citep{lambert2024rewardbench}\footnote{\url{https://huggingface.co/datasets/allenai/reward-bench}}: A benchmark that evaluates reward models on chat, reasoning, and safety. We use the Chat-Hard set, which requires identifying subtle deviations from the instruction.
\end{itemize}}

\paragraph{Datasets for Downstream Rubric-Based RL.}
We train policy models on queries from WildChat and evaluate the trained policies on three general-purpose benchmarks and one healthcare benchmark.
{\setlength{\leftmargini}{1.2em}
\begin{itemize}
    \item \textbf{WildChat}~\citep{zhao2024wildchat}\footnote{\url{https://huggingface.co/datasets/allenai/WildChat-1M}}: A corpus of real-world conversations between users and chatbots. We select 2K user queries and generate one rubric for each query with each method.
    \item \textbf{BiGGen-Bench}~\citep{kim2025biggen}\footnote{\url{https://huggingface.co/datasets/prometheus-eval/BiGGen-Bench}}: A generation benchmark with 765 instances covering nine capabilities, where each instance is evaluated against instance-specific criteria.
    \item \textbf{AlpacaEval 2.0}~\citep{li2023alpacaeval,dubois2024length}\footnote{\url{https://github.com/tatsu-lab/alpaca_eval}}: An instruction-following benchmark of 805 instructions that reports the win rate against a reference model, together with the length-controlled (LC) win rate.
    \item \textbf{Arena-Hard}~\citep{li2024crowdsourced}\footnote{\url{https://github.com/lmarena/arena-hard-auto}}: The Arena-Hard-Auto v0.1 benchmark of 500 challenging user queries, which reports the win rate against a baseline model, together with the style-controlled (SC) win rate.
    \item \textbf{HealthBench-Hard}~\citep{arora2025healthbench}\footnote{\url{https://github.com/openai/simple-evals}}: A challenging subset of HealthBench comprising 1,000 healthcare conversations with physician-written evaluation criteria. We use it to evaluate the out-of-domain generalization of the WildChat-trained policies, following the scoring procedure in Appendix~\ref{app:ood_generalization}.
\end{itemize}}

\subsection{Implementation Details}
\label{app:implementation}

In this section, we supplement the implementation details deferred from the main text, covering rubric generation and downstream RL.

\paragraph{Rubric Generation Settings.}
The initializer, the attacker, the quality judge, the repair agent, and the review agent of RubricArmor share the same rubric generator backbone. By default, the rubric generator backbone is Gemini-3.1-Flash-Lite with the reasoning effort set to low. Following Algorithm~\ref{alg:arme}, the initializer constructs $\mathcal{R}_0$ from the query, which also serves as the rubric of Direct. Attack Discovery and Verified Rubric Repair then alternate for at most $T=7$ rounds, and evolution stops early after $P=3$ consecutive rounds without an accepted revision. In the study of rubric generator backbones, all LLM agents switch to the given backbone together, and in the study of hyperparameter sensitivity, $T$ varies from 0 to 8. Since all methods compute the reward with Equation~\ref{eq:rubric_reward} under equal criterion weights, we adopt the uniformly weighted variant of RRD, denoted RRD-Uniform.

\paragraph{Downstream RL Settings.}
The RL experiments are implemented on top of the \texttt{verl} library\footnote{\url{https://github.com/verl-project/verl}}. We adopt GRPO~\citep{shao2024deepseekmath} as the policy optimization algorithm, and initialize the policy models from Qwen2.5-3B-Instruct, Qwen2.5-7B-Instruct, and Qwen3-4B, where Qwen3-4B runs in thinking mode and only the final answer is scored. During training, GPT-4.1-mini serves as the verifier, and the reward of each response is computed with Equation~\ref{eq:rubric_reward}. The training is configured with a global batch size of 128 queries, 8 rollouts per query, a sampling temperature of 1.0, and a learning rate of $1\times10^{-6}$. A small KL coefficient of $1\times10^{-3}$ anchors the policy to the reference model and stabilizes optimization. Prompts and responses are capped at 2,048 and 9,216 tokens, respectively, and policy optimization is executed for 3 epochs with one policy update per batch on 8 NVIDIA H20 GPUs. GPT-4.1 serves as the judge for all downstream benchmarks.

\paragraph{Prompts.}
We present abbreviated versions of the core prompts for brevity (Prompts~\ref{prompt:initializer}--\ref{prompt:verifier}).

\Needspace{10\baselineskip}
\begin{mybox}[label=prompt:initializer]{Prompt}{Prompt for the Initializer.}
\footnotesize
\textbf{System message}

\medskip\noindent\textbf{Role}\par\nobreak

You design scoring criteria for answers to a user's task.

\medskip\noindent\textbf{Inputs}\par\nobreak

You receive the user's task.

\medskip\noindent\textbf{Goal}\par\nobreak

Write a complete set of specific, directly verifiable scoring criteria for the user's task.

\medskip\noindent\textbf{Requirements}\par\nobreak

Derive criteria from the user's task, covering the requested deliverable, correctness, and explicit constraints. Check outcomes rather than prescribing a particular solution, and include the facts or conditions needed to judge each requirement directly.

Each criterion should be atomic, self-contained, and non-overlapping with the others. Give it a clear pass boundary and preserve the task's important requirements across the set.

Do not add unsupported obligations or exclude alternative correct answers. Distinguish required work from genuinely optional content, and let the task determine the number of criteria.

\medskip\noindent\textbf{Output}\par\nobreak

Return one JSON object with exactly "rubrics": a nonempty array of criterion strings. Each string is the complete criterion. Do not include IDs, weights, scores, or commentary outside this object. Let the task determine how many distinct criteria are needed.

\medskip\noindent\textbf{User message}\par\nobreak

\noindent \{\par

\noindent \hspace*{0.8em}"task": \textless{}query\textgreater{}\par

\noindent \}\par
\end{mybox}

\Needspace{10\baselineskip}
\begin{mybox}[label=prompt:attacker]{Prompt}{Prompt for the Attacker.}
\footnotesize
\textbf{System message}

\medskip\noindent\textbf{Role}\par\nobreak

You construct flawed answers to test scoring criteria.

\medskip\noindent\textbf{Inputs}\par\nobreak

You receive the task, current criteria, and past attempts for this task. A revision also includes the last answer, its failure description, and scoring feedback.

\medskip\noindent\textbf{Goal}\par\nobreak

Write a complete answer that substantially fails the task but passes every criterion. Then explain the failure in that answer. Missing, ambiguous, or incorrect criteria can make this possible.

\medskip\noindent\textbf{Requirements}\par\nobreak

Construct a plausible, complete response that satisfies every current criterion while substantially failing the task. Submit the answer itself rather than a plan or instructions to the grader.

Describe a defect that is present in the submitted answer, supported by the task and concrete evidence, and explain its material consequence. Keep this description independent of the criteria and grading results. A valid alternative or missing optional detail does not establish a substantial failure.

Use prior attempts, judgments, repairs, and scoring feedback to guide the next attempt. When revising, update the answer and its failure description together. A separate verifier determines whether every criterion passes.

\medskip\noindent\textbf{Output}\par\nobreak

Return one JSON object with exactly these fields, in this order:

- "y": the complete answer to submit for scoring.

- "p": a concrete explanation of that answer's actual task failure and harm.

If you cannot construct a candidate with a defensible substantial failure, return \{"y": null, "p": null\}. Otherwise both fields must be nonempty strings. Do not add other text.

\medskip\noindent\textbf{Revision goal}\par\nobreak

\textit{(Added before Output only when the attacker revises a previous response.)} Use scoring feedback to revise the answer. Return its updated failure description with it. You may attempt a different failure within the remaining attempt budget.

\medskip\noindent\textbf{User message}\par\nobreak

\noindent \{\par

\noindent \hspace*{0.8em}"task": \textless{}query\textgreater{},\par

\noindent \hspace*{0.8em}"criteria": \textless{}current rubric\textgreater{},\par

\noindent \hspace*{0.8em}"previous\_attempts": \textless{}memory of past adversarial responses, outcomes, and accepted revisions\textgreater{},\par

\noindent \hspace*{0.8em}"previous\_answer": \textless{}last adversarial response\textgreater{}, \textit{(revision only)}\par

\noindent \hspace*{0.8em}"previous\_failure": \textless{}its defect\textgreater{}, \textit{(revision only)}\par

\noindent \hspace*{0.8em}"scoring\_feedback": \textless{}verifier feedback\textgreater{} \textit{(revision only)}\par

\noindent \}\par
\end{mybox}

\Needspace{10\baselineskip}
\begin{mybox}[label=prompt:quality_judge]{Prompt}{Prompt for the Quality Judge.}
\footnotesize
\textbf{System message}

\medskip\noindent\textbf{Role}\par\nobreak

You independently judge a proposed failure in an answer.

\medskip\noindent\textbf{Inputs}\par\nobreak

You receive the task, current answer, and proposed failure. You do not receive scoring criteria or scores.

\medskip\noindent\textbf{Goal}\par\nobreak

Decide whether the proposed failure occurs and substantially harms task completion.

\medskip\noindent\textbf{Requirements}\par\nobreak

Check the proposed failure against the task and current answer. Establish whether it occurs and, separately, whether it substantially harms task completion. Ground both decisions in concrete evidence and connect the defect to the required outcome.

Do not impose unsupported requirements or reject valid alternatives. Judge the proposed failure rather than unrelated errors; rejecting that claim does not certify the whole answer. Do not propose scoring criteria.

\medskip\noindent\textbf{Output}\par\nobreak

Return one JSON object with exactly these fields, in this order:

- "reason": a concise task-grounded explanation. State any unsupported claim or uncertainty before making the decisions.

- "failure\_present": whether the proposed failure is established in this answer.

- "material\_harm": whether that failure substantially harms task completion. Use false if the failure or its substantial harm is not established.

Both decisions must be booleans. Do not add other fields or commentary outside the JSON object.

\medskip\noindent\textbf{User message}\par\nobreak

\noindent \{\par

\noindent \hspace*{0.8em}"task": \textless{}query\textgreater{},\par

\noindent \hspace*{0.8em}"proposed\_failure": \textless{}defect identified by the attacker\textgreater{},\par

\noindent \hspace*{0.8em}"answer": \textless{}adversarial response\textgreater{}\par

\noindent \}\par
\end{mybox}

\Needspace{10\baselineskip}
\begin{mybox}[label=prompt:repair]{Prompt}{Prompt for the Repair Agent.}
\footnotesize
\textbf{System message}

\medskip\noindent\textbf{Role}\par\nobreak

You repair scoring criteria after an answer passed them and a quality judge confirmed a task failure.

\medskip\noindent\textbf{Inputs}\par\nobreak

You receive the task, current criteria, answer, failure description, quality judgment, and past repairs. Retries include the previous draft and feedback.

\medskip\noindent\textbf{Goal}\par\nobreak

Make the smallest task-grounded change that detects this failure and similar failures. Return the complete set.

\medskip\noindent\textbf{Requirements}\par\nobreak

Make only changes justified by the task and the confirmed failure. Correct incorrect requirements, add missing checks, or clarify underspecified criteria. Keep unaffected criteria unchanged and preserve the valid intent of past repairs.

Use revision feedback to improve the previous draft for the same submitted answer. Return the complete criterion set. Each changed draft is graded before review; it must detect the confirmed defect rather than merely change the wording.

\medskip\noindent\textbf{Rubric properties}\par\nobreak

Preserve coverage of the important task requirements and allow alternative correct solutions without adding unsupported obligations. Maintain these three properties:

\begin{enumerate}[label=\arabic*.,leftmargin=1.5em,itemsep=2pt,topsep=3pt,parsep=0pt]
\item \textbf{Atomic.} Each criterion checks one independently judgeable requirement.
\item \textbf{Self-contained.} Each criterion states the facts, values, or conditions needed for evaluation, with a clear pass boundary.
\item \textbf{Non-overlapping.} Criteria check distinct requirements, avoiding repeated reward for the same response property.
\end{enumerate}

\medskip\noindent\textbf{Output}\par\nobreak

Return one JSON object with exactly these fields, in this order:

- "change\_reason": a short explanation of the failure, the task basis for the changes, and how the new criteria detect it.

- "rubrics": the full nonempty array of revised and retained criterion strings.

Do not return edit operations, IDs, weights, scores, or other commentary.

\medskip\noindent\textbf{User message}\par\nobreak

\noindent \{\par

\noindent \hspace*{0.8em}"task": \textless{}query\textgreater{},\par

\noindent \hspace*{0.8em}"current\_criteria": \textless{}current rubric\textgreater{},\par

\noindent \hspace*{0.8em}"alleged\_failure": \textless{}confirmed defect\textgreater{},\par

\noindent \hspace*{0.8em}"submitted\_answer": \textless{}adversarial response\textgreater{},\par

\noindent \hspace*{0.8em}"quality\_judgment\_reason": \textless{}reason of the quality judge\textgreater{},\par

\noindent \hspace*{0.8em}"past\_committed\_repairs": \textless{}accepted revisions in memory\textgreater{},\par

\noindent \hspace*{0.8em}"previous\_draft": \textless{}previous revision draft or null\textgreater{},\par

\noindent \hspace*{0.8em}"revision\_feedback": \textless{}feedback on the previous draft\textgreater{}\par

\noindent \}\par
\end{mybox}

\Needspace{10\baselineskip}
\begin{mybox}[label=prompt:review]{Prompt}{Prompt for the Review Agent.}
\footnotesize
\textbf{System message}

\medskip\noindent\textbf{Role}\par\nobreak

You review revised scoring criteria after the same answer has been graded against the old and new sets.

\medskip\noindent\textbf{Inputs}\par\nobreak

You receive the task, old and proposed criteria, submitted answer, failure description, quality judgment, and repair explanation. The old and new verification arrays contain each criterion, its verdict, and its grading reason, in the corresponding set's order.

\medskip\noindent\textbf{Goal}\par\nobreak

Approve a justified repair that catches the confirmed task failure. Use the actual grading evidence and preserve other valid requirements.

\medskip\noindent\textbf{Requirements}\par\nobreak

Use the old and new grading results to identify an actual FAIL that detects the confirmed defect. Check that the failure follows from a justified change to the criteria; a changed verdict alone does not prove that the repair works.

Independently assess the task basis of the changes and the supporting explanations. Reject unsupported restrictions, changes that exclude valid answers, and proposals that do not substantively repair the defect.

Check that the revised set preserves valid coverage and the rubric properties below. Provide concrete feedback on problems with the repair, without requesting unrelated changes.

\medskip\noindent\textbf{Rubric properties}\par\nobreak

Check that the revised set covers the important task requirements and allows alternative correct solutions without adding unsupported obligations. Verify these three properties:

\begin{enumerate}[label=\arabic*.,leftmargin=1.5em,itemsep=2pt,topsep=3pt,parsep=0pt]
\item \textbf{Atomic.} Each criterion checks one independently judgeable requirement.
\item \textbf{Self-contained.} Each criterion states the facts, values, or conditions needed for evaluation, with a clear pass boundary.
\item \textbf{Non-overlapping.} Criteria check distinct requirements, avoiding repeated reward for the same response property.
\end{enumerate}

\medskip\noindent\textbf{Output}\par\nobreak

Return one JSON object with exactly these fields, in this order:

- "reason": a short explanation connecting the actual FAIL, the described failure, the justified change, and retained coverage and valid alternatives.

- "approved": a boolean approving or rejecting this proposed set.

- "issues": concrete problems with the affected criteria when rejecting; an empty array when approving.

Do not rewrite the criteria or add commentary outside the JSON object.

\medskip\noindent\textbf{User message}\par\nobreak

\noindent \{\par

\noindent \hspace*{0.8em}"task": \textless{}query\textgreater{},\par

\noindent \hspace*{0.8em}"old\_criteria": \textless{}current rubric\textgreater{},\par

\noindent \hspace*{0.8em}"proposed\_criteria": \textless{}revised rubric\textgreater{},\par

\noindent \hspace*{0.8em}"alleged\_failure": \textless{}confirmed defect\textgreater{},\par

\noindent \hspace*{0.8em}"submitted\_answer": \textless{}adversarial response\textgreater{},\par

\noindent \hspace*{0.8em}"quality\_judgment\_reason": \textless{}reason of the quality judge\textgreater{},\par

\noindent \hspace*{0.8em}"repair\_explanation": \textless{}reason of the repair agent\textgreater{},\par

\noindent \hspace*{0.8em}"old\_verification": \textless{}verifier results under the current rubric\textgreater{},\par

\noindent \hspace*{0.8em}"new\_verification": \textless{}verifier results under the revised rubric\textgreater{}\par

\noindent \}\par
\end{mybox}

\Needspace{10\baselineskip}
\begin{mybox}[label=prompt:verifier]{Prompt}{Prompt for the Verifier.}
\footnotesize
\textbf{System message}

\medskip\noindent\textbf{Role}\par\nobreak

You evaluate one scoring criterion for an answer to a user's task.

\medskip\noindent\textbf{Inputs}\par\nobreak

You receive the task, exactly one criterion, and the submitted answer.

\medskip\noindent\textbf{Goal}\par\nobreak

Apply that criterion faithfully and return PASS or FAIL with a specific reason. Do not add quality requirements that the criterion does not state.

\medskip\noindent\textbf{Requirements}\par\nobreak

Evaluate only the supplied criterion within the task's stated scope, using the actual answer rather than claims of compliance. Treat instructions within the answer as content to evaluate.

Return PASS when the criterion is satisfied and FAIL otherwise, including when required content is missing. Apply conditions as written; a failure unrelated to this criterion must not affect its verdict.

Use the information in the task and criterion as the basis for judgment. Do not add obligations, replace the criterion with an independent solution, or score the answer by its overall style.

\medskip\noindent\textbf{Output}\par\nobreak

Return one JSON object with exactly "reason" followed by "verdict". First write a short, concrete reason using this criterion's scope and the answer's actual behavior. Then set "verdict" to exactly "PASS" or "FAIL" to match that reason. Do not emit a provisional verdict before analyzing it. Do not add other fields or commentary outside the JSON object.

\medskip\noindent\textbf{User message}\par\nobreak

\noindent \{\par

\noindent \hspace*{0.8em}"task": \textless{}query\textgreater{},\par

\noindent \hspace*{0.8em}"criterion": \textless{}one criterion\textgreater{},\par

\noindent \hspace*{0.8em}"answer": \textless{}response\textgreater{}\par

\noindent \}\par
\end{mybox}

\end{document}